\documentclass[11pt]{article}

\usepackage[final]{acl}

\usepackage{times}
\usepackage{latexsym}
\usepackage{amsmath}

\usepackage[T1]{fontenc}

\usepackage[utf8]{inputenc}

\usepackage{microtype}

\usepackage{inconsolata}

\usepackage{graphicx}
\usepackage{enumitem}

\usepackage{graphicx}
\usepackage{enumitem}
\usepackage{booktabs}
\usepackage{tabularx}
\usepackage{array}
\usepackage{multirow}
\usepackage{url}

\newcolumntype{Y}{>{\raggedright\arraybackslash}X}
\newcolumntype{L}[1]{>{\raggedright\arraybackslash}p{#1}}
\newcommand{\benchmark}{Evidence Package Benchmark}
\newcommand{\sys}{EviBound}
\definecolor{msblue}{RGB}{235,242,250}
\definecolor{mspeach}{RGB}{252,239,224}
\definecolor{msgray}{RGB}{242,242,242}
\definecolor{msline}{RGB}{170,170,170}
\setlist[itemize]{leftmargin=*,nosep}
\author{
  Chengyuan Gao \textsuperscript{1,2,3} \quad
  Jiang Wu \textsuperscript{4} \quad
  Tao Lu \textsuperscript{3} \quad
  Jiayan Guo \textsuperscript{5} \\
  {\bf Mingkun Xu}\textsuperscript{\textbf{4}} \quad
  {\bf Tianyi Zang}\textsuperscript{\textbf{2*}} \quad
  {\bf Shangyang Li}\textsuperscript{\textbf{1,3*}} \\
  \textsuperscript{1}Beijing University of Posts and Telecommunications \quad
  \textsuperscript{2}Harbin Institute of Technology \\
  \textsuperscript{3}Renyixun Health Technology Co., Ltd. \quad
  \textsuperscript{4}GDIIST \quad
  \textsuperscript{5}Tencent \\
  \texttt{\{1967464882guge, wuj2386, taolu5010\}@gmail.com} \quad
  \texttt{guojy001@outlook.com} \\
  \texttt{xumingkun@gdiist.cn} \quad
  \texttt{tianyi.zang@hit.edu.cn} \quad
  \texttt{syli@bupt.edu.cn} \\
}

\newcommand\blfootnote[1]{%
  \begingroup
  \renewcommand\thefootnote{}\footnote{#1}%
  \addtocounter{footnote}{-1}%
  \endgroup
}

\title{Evidence-Bounded Mental Health Reasoning from Heterogeneous Speech Protocols}

\begin{document}
\maketitle

\blfootnote{$^*$Corresponding author.}

\begin{abstract}
Computational mental health screening using multimodal speech and text has shown great promise. However, existing models often assume all clinical speech protocols carry equivalent evidentiary validity. In reality, heterogeneous protocols, from free interviews to fixed reading tasks, support fundamentally different evidence. Forcing uniform reasoning flattens these boundaries, causing models to hallucinate symptoms from irrelevant text or overclaim support. Even advanced long chain-of-thought LLMs fail to resolve this issue, as free-form reasoning can exacerbate boundary violations.  
To address this, we reformulate multimodal screening as an evidence-bounded reasoning problem. We introduce the \textbf{Evidence Package Benchmark}, integrating 1,870 packages across six heterogeneous sources with explicit modality masks and evidence permissions. We further propose \textbf{EviBound}, a protocol-aware evidence control framework. Unlike direct LLM prompting, EviBound uses a profile-aware planner to restrict reasoning scope, orchestrates evidence tools via five-way acoustic consensus, and enforces a boundary critic to suppress unsupported claims.  
Empirical results show EviBound achieves a held-out test Depression AUROC of 0.8658, exceeding the strongest direct omni-modal baseline by +0.0811 AUROC while maintaining zero claim violations. Our work moves beyond unconstrained accuracy toward evidence-consistent, protocol-aware systems for safer clinical NLP research. Code will be released upon publication. 
\end{abstract}

\section{Introduction}
\begin{figure}[t]
\centering
\includegraphics[width=0.99\linewidth]{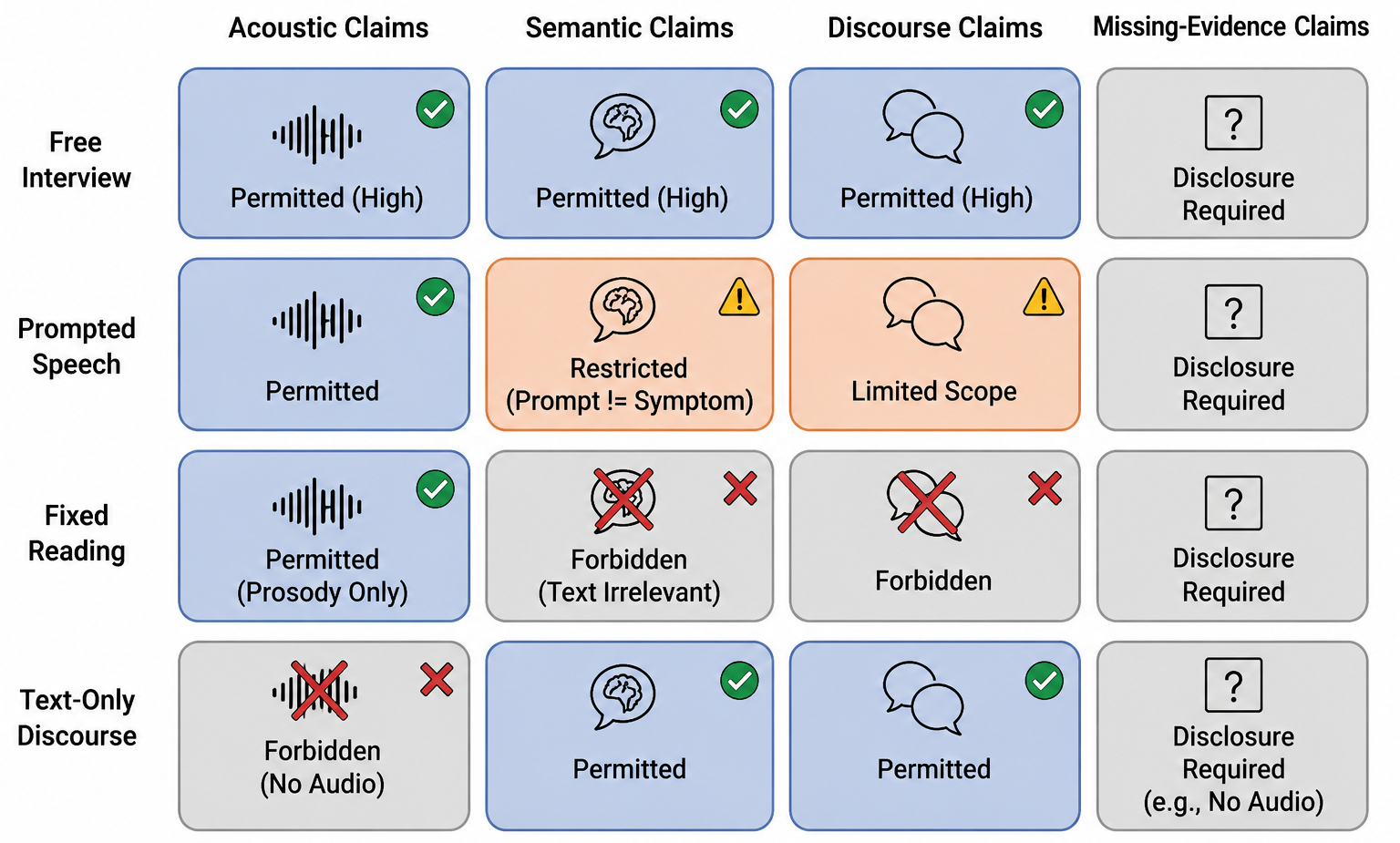}
\caption{Protocol-specific evidence boundaries across heterogeneous speech records. Free interviews, prompted speech, fixed reading, and text-only discourse license different levels of acoustic, semantic, discourse, and missing-evidence claims.}
\label{fig:protocol-boundaries}
\end{figure}

Large language models (LLMs) offer unprecedented opportunities for scalable mental health screening. However, contemporary omni-modal systems often implicitly assume that \emph{all speech acquisition protocols provide equivalent evidentiary validity}. In practice, diagnostic evidence is inherently constrained by its collection context. As illustrated in Figure~\ref{fig:protocol-boundaries}, mental health datasets follow four protocol types: free interviews (supporting acoustic, semantic, and discourse claims), prompted speech (acoustic evidence permitted but semantic inference restricted), fixed reading (prosody only, text irrelevant), and text-only discourse (no prosody, semantic claims from language alone)\cite{daic_woz,avec2019,cmdc}. Each protocol thus establishes distinct boundaries for acoustic, semantic, discourse, and missing-evidence claims~\cite{qin2025mental,eatd_corpus,modma,discourse_uwo,dais_c}.

Ignoring these boundaries leads to what we term \emph{epistemic flattening}: the collapse of heterogeneous evidence constraints into a single unconstrained reasoning space~\cite{qin2025mental,wang2026emotionthinker}. Under restrictive protocols, systems hallucinate symptoms by conflating semantic and acoustic cues, generate unsupported interpretations, and attribute conclusions to absent modalities. Importantly, neither larger models nor longer chain-of-thought reasoning resolves this issue; they can instead increase reasoning instability and computational cost while exacerbating boundary violations~\cite{liu2026benchmarking,more_thinking_less_seeing,mirage,mme_cot}. These findings indicate that the central challenge is not insufficient reasoning capability, but the absence of explicit evidence-bound control.

To address this limitation, we reformulate computational mental health screening as an evidence-bounded reasoning problem. Formally, we define each screening instance as an \emph{evidence package} $\mathcal{E} = (P, M, \Pi, \mathcal{X})$, where $P$ is the protocol profile, $M$ the modality mask, $\mathcal{X}$ the multimodal observations, and $\Pi$ the admissible evidence boundary for that protocol. Under this formulation, the objective is not only to maximize predictive accuracy, but also to ensure conclusions remain strictly consistent with observable evidence and protocol constraints. This view turns dataset documentation, model reporting, and medical/psychiatric benchmark principles into executable constraints rather than prose-only caveats~\cite{datasheets,model_cards,healthbench}. Importantly, this task is framed as benchmark-level boundary control for research evaluation rather than formal clinical diagnosis or deployment readiness~\cite{datasheets,model_cards,healthbench}.

To operationalize this paradigm, we introduce \textbf{EviBound}, a protocol-aware evidence control framework for bounded multimodal reasoning. Instead of unrestricted direct generation, EviBound uses a three-stage architecture: (1) a profile-aware planner that defines permissible reasoning scope based on protocol and modality availability; (2) parallel evidence modules that cross-validate signals via five-way acoustic consensus; and (3) a boundary critic that suppresses unsupported claims and enforces consistency with the modality mask.

Our primary contributions are:
\begin{itemize}[noitemsep,topsep=2pt,leftmargin=*,itemindent=0pt]
    \item \textbf{Evidence-Bounded Reasoning.} We formalize protocol-aware mental health reasoning with explicit evidence constraints under heterogeneous speech settings.
    
    \item \textbf{Evidence Package Benchmark.} We build a benchmark of 1,870 standardized evidence packages from six clinical sources with modality masks and claim permissions.
    
    \item \textbf{EviBound Framework.} We propose a protocol-aware multimodal reasoning harness with explicit evidence routing and boundary validation.
    
    \item \textbf{Boundary-Aware Evaluation.} We introduce CVR, MHR, and EBCPass to measure evidence consistency under heterogeneous protocols.
    
    \item \textbf{Experimental Results.} EviBound achieves 0.8658 AUROC and 100\% EBCPass on held-out packages with zero boundary violations.
\end{itemize}

\section{Related Work}
\label{sec:related_work}

\paragraph{Multimodal Mental Health Screening and Clinical Benchmarks.}
Speech-based mental health research spans interview modeling, multimodal fusion, audiovisual risk assessment, and recent large multimodal reasoning systems for psychiatric analysis~\cite{al2018detecting,fang2023multimodal,dhelim2023artificial,zhang2026mllm,qin2025mental,wang2026emotionthinker}. Public resources such as DAIC-WOZ/E-DAIC, CMDC, EATD, MODMA, MMPsy, DISCOURSE-UWO, and DAIS-C differ substantially in language, modality availability, and acquisition protocol~\cite{daic_woz,avec2019,cmdc,eatd_corpus,modma,discourse_uwo,dais_c}. Meanwhile, benchmarks including HealthBench, MedHELM, MentalBench, and PsychiatryBench emphasize that clinical AI evaluation should extend beyond a single predictive metric toward reliability and reasoning assessment~\cite{healthbench,jing2026beyond,mentalbench,psychiatrybench}. However, prior systems largely treat heterogeneous protocols as evidentially interchangeable, often producing what we term \emph{epistemic flattening}. In contrast, our benchmark explicitly models modality masks and claim permissions as reasoning constraints.

\paragraph{Speech Evidence, Clinical Agents, and Trustworthy Medical AI.}
Recent work converts acoustic landmarks, SSL representations, digital phenotypes, and psychological knowledge into LLM-usable evidence for mental-health prediction~\cite{acoustic_landmarks,speechtrag,llm_dep_psych_knowledge,speech_digital_phenotype,cfrafn}. Standard speech pipelines rely on openSMILE/eGeMAPS and SSL encoders such as wav2vec 2.0, HuBERT, and WavLM~\cite{opensmile,wav2vec2,hubert,wavlm}, while clinical language-agent systems increasingly employ structured planning and tool orchestration for biomedical reasoning~\cite{tang2024medagents,yue2024clinicalagent,wang2025medagentproevidencebasedmultimodalmedical,liu2026benchmarking, mao2026schemaadaptivetabularrepresentationlearning, zhi2026medgr2, zheng-etal-2026-evo}. Concurrently, trustworthy clinical NLP studies investigate hallucination mitigation, verifier-guided decoding, retrieval-augmented generation, and factuality-oriented evaluation~\cite{ni2025trustworthyretrievalaugmentedgeneration,Han2025VerifiAgentAU}. Unlike prior work that primarily expands reasoning capability or factual grounding, our focus is protocol-aware evidence validity through an Agent-Evidence Interface and boundary-aware evaluation metrics (CVR, MHR, and EBCPass).

\section{Methodology}
\label{sec:method}

\begin{figure*}[t]
\centering
\includegraphics[width=0.99\linewidth]{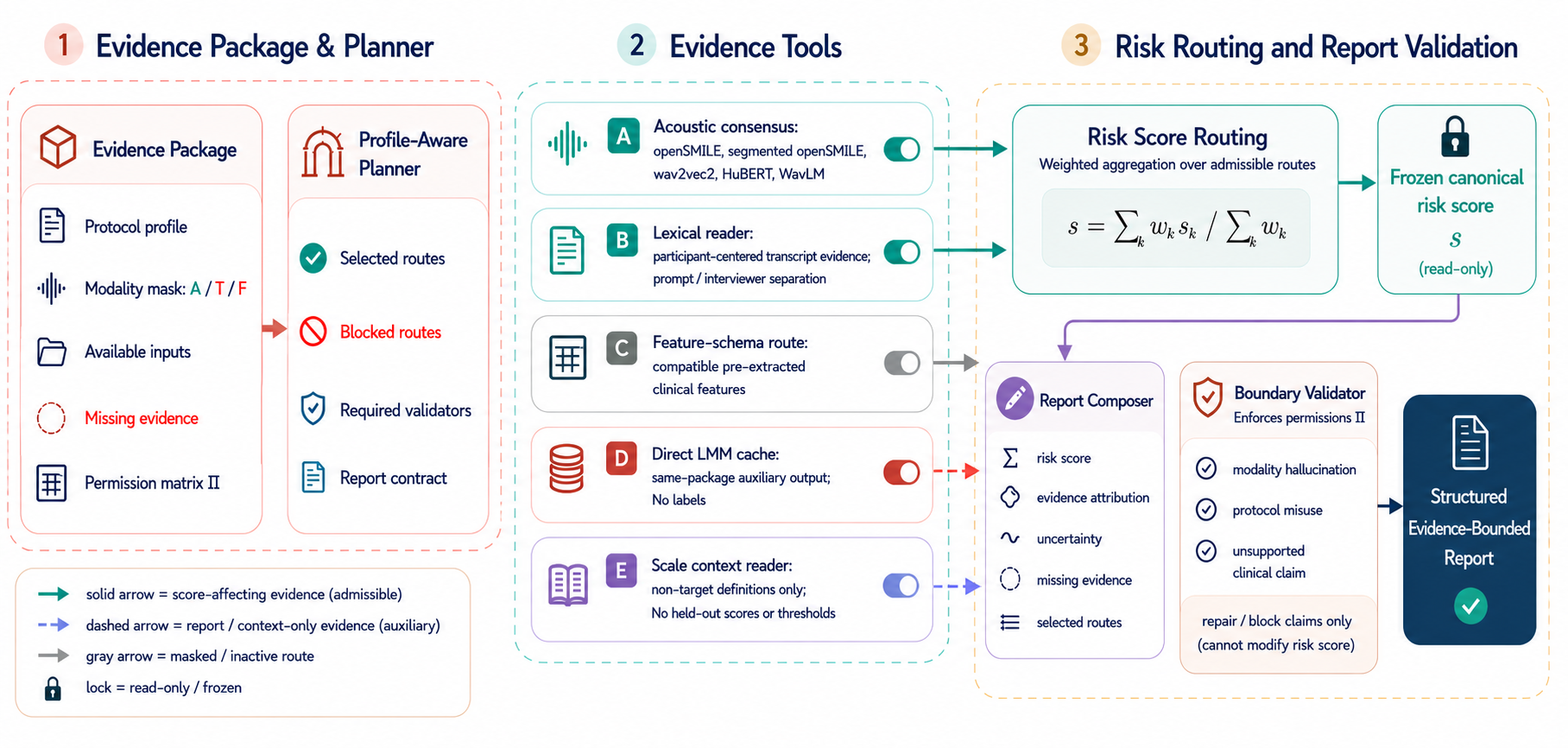} 
\caption{The \sys{} framework architecture. (1) The Profile-Aware Planner parses the evidence package $\mathcal{E}$ to select admissible routes and enforce the permission matrix $\Pi$. (2) Evidence Tools extract modality-specific signals, including a five-way acoustic consensus and lexical/feature readers. (3) The system decouples Risk Score Routing from Report Validation: risk scores are aggregated from admissible routes and frozen, while a deterministic validator ensures the final structured report respects evidence boundaries without altering the read-only risk score.}
\label{fig:framework}
\end{figure*}

We formalize \emph{evidence-bounded mental health reasoning} and present \textbf{EviBound}, a protocol-aware structured evidence harness for multimodal screening under heterogeneous acquisition protocols. Unlike conventional multimodal systems that implicitly treat all modalities as evidentially interchangeable, EviBound operates over explicit \emph{evidence packages} encoding protocol profile, modality availability, missing evidence, and admissible claim scope. The framework separates three components: (1) an \emph{evidence interface} defining protocol-conditioned routing constraints; (2) a \emph{risk routing layer} that aggregates compatible acoustic, semantic, and feature evidence; and (3) a \emph{boundary validator} that enforces evidence consistency during report generation.

\subsection{Evidence-Bounded Reasoning}

A central but largely unexamined assumption in multimodal mental health screening is that all speech protocols support equivalent forms of clinical inference. In practice, fixed-reading audio, prompted narratives, open-ended interviews, and text-only discourse expose fundamentally different evidence signals. Ignoring these distinctions produces what we term \emph{epistemic flattening}: heterogeneous evidence constraints collapse into a single unconstrained reasoning space, allowing systems to infer symptoms from scripted text or generate acoustic claims without audio evidence.

We therefore formulate screening as an \emph{evidence-bounded reasoning} problem. Each instance is represented as an \textbf{evidence package}
\[
\mathcal{E}=(P,M,\Pi,\mathcal{X}),
\]
where $P$ denotes the protocol profile, $M$ the modality mask, $\mathcal{X}$ the observed multimodal inputs, and $\Pi$ the admissible evidence boundary defined by $P$ and $M$. Protocol profiles distinguish acquisition settings such as interviews, prompted speech, fixed reading, and text-only discourse, while modality masks specify available evidence channels including raw audio (A), transcript text (T), and pre-extracted acoustic features (F).

Given an evidence package $\mathcal{E}$, the system predicts screening-oriented risk scores and generates a structured report $r$. A report is considered \emph{evidence-bounded} only if every emitted claim remains consistent with the package profile and modality permissions:
\[
\mathrm{permit}(c,M,P)=1,\quad \forall c\in\mathcal{C}(r),
\]
where $\mathcal{C}(r)$ denotes the set of generated claims. The permission function is protocol-dependent: direct listening claims require raw audio, symptom-history claims require participant-centered interview evidence, and fixed-reading text cannot support semantic symptom inference. Missing evidence must also be explicitly disclosed.

The objective is therefore not only predictive performance, but also evidence-consistent reasoning under heterogeneous acquisition protocols.

\subsection{Evidence Package Benchmark}

We construct the \textbf{Evidence Package Benchmark}, a unified benchmark of 1,870 packages derived from six heterogeneous mental health resources: CMDC, DAIS-C, E-DAIC, EATD, MMPsy, and MODMA. Rather than concatenating raw datasets into a homogeneous cohort, each source is instantiated into the unified evidence package schema $\mathcal{E}=(P,M,\Pi,\mathcal{X})$, where observations are strictly partitioned into \emph{model-visible inputs} and \emph{evaluator-only labels} $y_{\mathrm{eval}}$ (e.g., PHQ-9 or MDD diagnosis). Evaluator-side labels remain inaccessible during inference, preventing leakage through scale memorization or protocol shortcuts.

Table~\ref{tab:datasets} summarizes source composition, available modalities, languages, and frozen subject-independent splits. Packages may contain audio, transcript text, feature vectors, or text-only discourse. Feature-only records can support feature-derived risk estimation but cannot license direct listening claims. Depression screening, anxiety screening, discourse probes, and severity-oriented tasks remain distinct evaluation surfaces rather than being collapsed into a universal target.

\begin{table}[t]
\centering
\scriptsize
\setlength{\tabcolsep}{0.70pt}
\begin{tabularx}{\linewidth}{@{}l c r c Y c Y@{}}
\toprule
Data & Lang. & $N$ & Split & Profile & Inputs & Eval-only labels \\
\midrule
CMDC & Zh & 78 & 57/8/13 & interview & A+T & MDD/HC; PHQ/HAMD \\
DAIS-C & En & 28 & 18/4/6 & discourse & T & group/disc. probe \\
E-DAIC & En & 275 & 163/56/56 & interview & A+T+F & PHQ-8 dep.; S \\
EATD & Zh & 162 & 116/13/33 & prompted & A & SDS dep.; S \\
MMPsy & Zh & 1275 & 891/128/256 & interview & T+F & PHQ/GAD; S \\
MODMA & Zh & 52 & 39/5/8 & reading & A & MDD/HC; PHQ/GAD \\
\bottomrule
\end{tabularx}
\caption{Benchmark composition with fixed package splits. $N$ counts packages, not raw segments. Inputs are model-visible A/T/F fields; labels and scale-derived targets (S) are evaluator-only.}
\label{tab:datasets}
\end{table}

The claim-permission matrix (Table~\ref{tab:permission}) defines the evidential scope associated with each protocol--modality combination and serves as the executable specification for runtime report validation.

\begin{table}[t]
\centering
\scriptsize
\setlength{\tabcolsep}{1.6pt}
\begin{tabularx}{\linewidth}{@{}l c c Y@{}}
\toprule
Case & Voice claim & Text claim & Main restriction \\
\midrule
Interview + A & direct & yes & split participant/prompt \\
Interview no A & F-derived & yes & no direct listening \\
Prompted + A & acoustic & limited & do not treat prompt text as symptoms \\
Reading + A & acoustic & no & no personal-history inference \\
Text/disc. & no & text/disc. & no acoustic claims \\
\bottomrule
\end{tabularx}
\caption{Compact claim-permission matrix. Rows are profile/mask cases. A=raw audio, T=transcript, F=pre-extracted features.}
\label{tab:permission}
\end{table}

\subsection{Profile-Aware Planner}

EviBound uses a deterministic planner to map each evidence package to a restricted set of compatible evidence routes. The planner functions as an \emph{Agent--Evidence Interface}, determining which tools, claims, and report fields are admissible before reasoning begins.

Clinical interviews may activate lexical, acoustic, and feature routes; prompted or fixed-reading speech may use acoustic evidence but cannot interpret scripted text as symptom history; text-only records may support discourse analysis but cannot generate acoustic or prosodic claims. Missing modalities are explicitly registered before inference, ensuring that absent evidence becomes part of the report contract rather than an implicit assumption.

\subsection{Evidence Tools and Risk Routing}

The evidence harness integrates lightweight and independently auditable evidence modules. For audio-supported records, the primary route is a five-way acoustic consensus over openSMILE/eGeMAPS, segmented openSMILE, wav2vec2, HuBERT, and WavLM representations. Each branch outputs calibrated risk estimates and uncertainty signals, enabling conservative aggregation under disagreement.

For the applicable route set $R(x)=\{k:\rho_k(x)=1\}$, where $\rho_k$ is the route precondition determined by protocol and modality permissions, the final routed score is computed as
\begin{equation}
\hat{s}(x)=
\frac{\sum_{k\in R(x)} w_k s_k(x)}
{\sum_{k\in R(x)} w_k},
\end{equation}
where weights $w_k \ge 0$ are historical calibration performance. Inapplicable routes are masked before aggregation, and disagreement signals propagate into report uncertainty fields.

Beyond acoustic routing, EviBound includes lexical readers for participant-centered transcripts, feature-schema routes for compatible pre-extracted clinical features, and frozen omni-modal LLM outputs used as auxiliary evidence sources rather than authoritative decision makers. Source identifiers are used only for compatibility masking and never as direct predictors of clinical status.

\subsection{Boundary Validator and Structured Reports}

The report composer emits a structured schema containing risk score, uncertainty, evidence attribution, missing evidence, blocked claims, selected routes, and validator status. The final risk score is frozen before report repair, ensuring that post-hoc validation cannot alter AUROC or F1 outcomes.

The validator operates as the executable realization of the package permission matrix, transforming protocol constraints into runtime-verifiable report boundaries. It checks three major categories of violations:
\begin{enumerate}[leftmargin=*, topsep=3pt, itemsep=2pt]
    \item \textbf{Modality hallucination}: claims referencing unavailable modalities are removed;
    \item \textbf{Protocol misuse}: fixed-reading or prompted text cannot support symptom-history inference;
    \item \textbf{Claim scope}: diagnosis, treatment, prognosis, or unsupported clinical recommendations are blocked.
\end{enumerate}

Unsupported claims are removed or rewritten, blocked reasons are recorded, and schema validation is rerun before output finalization. The validator operates under a strict read-only constraint on risk scores:
\begin{equation}
\operatorname{score}(r')=\operatorname{score}(r),
\end{equation}
where $r'=\operatorname{validate}(r;x)$ denotes the repaired report. Report validation is therefore evaluated independently from predictive classification performance.

\subsection{Inference and Evaluation}

EviBound operates as a compositional protocol-governed pipeline, as illustrated in Figure~\ref{fig:framework}. During inference, the planner first selects compatible routes from the evidence package, followed by parallel evidence extraction, structured report generation, and deterministic boundary validation. This design decouples multimodal evidence aggregation from report-boundary enforcement, reducing unsupported reasoning drift during generation.
\begin{equation}
\mathcal{E} \xrightarrow{\text{Planner}} \tau \xrightarrow{\text{Tools}} \{\mathcal{C}_a, X_s\} \xrightarrow{\text{Decoder}} \mathcal{R}
\end{equation}

We evaluate both predictive utility and evidence consistency. Predictive performance is measured using AUROC, Macro-$F_1$, and Quadratic Weighted Kappa (QWK) for severity-oriented evaluation. Evidence consistency is audited using Claim Violation Rate (CVR), Modality Hallucination Rate (MHR), and Evidence-Bound Consistency Pass (EBCPass), which measures whether generated claims remain fully contained within the admissible evidence boundary $\Pi$. Importantly, replay diagnostics and report-boundary auditing are evaluated independently from disease-label ranking metrics.

\section{Experimental Setup}

\subsection{Baselines}

We compare EviBound against four categories of baselines under the identical package-interface setting, ensuring that all systems operate only on model-visible evidence without access to evaluator-only labels or unavailable modalities.

\textbf{Direct omni-modal LMMs.}  
We evaluate direct prompting baselines using Qwen3-Omni-Flash, Qwen3.5-Omni-Plus, and Gemini 3.5 Flash~\citep{qwen3_omni,qwen35_omni,google2026gemini35flash}. These systems receive the same package inputs, structured prompts, decoding configuration, and output schema as EviBound, but generate reports without explicit evidence-bound routing or boundary validation.

\textbf{Long-reasoning baseline.}  
To test whether extended chain-of-thought reasoning alone can recover protocol-aware behavior, we evaluate a long-reasoning baseline with expanded reasoning budgets and structured deliberation prompts. This setting follows recent multimodal reasoning analyses showing that longer reasoning traces do not necessarily improve grounded multimodal inference~\citep{more_thinking_less_seeing,mirage,mme_cot}.

\textbf{Acoustic and feature baselines.}  
We include conventional speech-analysis baselines covering openSMILE/eGeMAPS descriptors, segmented acoustic aggregation, SSL embeddings (Wav2Vec2.0, HuBERT, WavLM), and the proposed five-way acoustic consensus route~\citep{opensmile,wav2vec2,hubert,wavlm}. We additionally evaluate feature-based logistic regression models over compatible pre-extracted clinical features, as well as LLM-facing speech-evidence pipelines that transform acoustic or psychological signals into language-readable evidence~\citep{acoustic_landmarks,speechtrag,llm_dep_psych_knowledge,speech_digital_phenotype,cfrafn}.

\textbf{Harness ablations.}  
To isolate the contribution of each component, we evaluate controlled ablations that separately remove profile-aware planning, acoustic consensus, feature routing, and boundary validation. Predictive routing components are evaluated independently from report-repair modules to avoid conflating score improvements with post-processing effects.

All systems share the same package ordering, parser implementation, decoding limits, cached outputs, and evaluation scripts.

\subsection{Metrics}

We evaluate models along two complementary dimensions: \emph{predictive performance} and \emph{evidence consistency}. This separation allows reasoning validity and predictive utility to be evaluated independently.

\paragraph{Predictive Metrics.}
Predictive evaluation uses held-out evidence packages only. We report AUROC and validation-threshold F1 on the pooled depression-eligible split ($n=366$, 94 positives) and the anxiety-eligible split ($n=264$, 35 positives, dominated by MMPsy). Severity-oriented evaluation further reports Quadratic Weighted Kappa (QWK) on compatible severity subsets ($n=320$). Secondary diagnostics include AUPRC, Brier score, and calibration-oriented operating points. Depression and anxiety evaluations are reported separately because protocol composition and class distributions differ substantially across tasks.

\paragraph{Evidence Consistency Metrics.}
Beyond predictive accuracy, we evaluate whether generated reports remain consistent with evidence permissions through protocol-aware replay diagnostics. We formalize three boundary-aware metrics:
\begin{itemize}[leftmargin=*, topsep=3pt, itemsep=2pt]
    \item \textbf{Claim Violation Rate (CVR):} proportion of packages containing at least one protocol-violating claim (e.g., no-audio acoustic hallucination or fixed-reading symptom overreach);
    \item \textbf{Missing-Evidence Handling Rate (MHR):} proportion of packages correctly disclosing unavailable modalities;
    \item \textbf{EBCPass (Evidence-Bound Consistency Pass):} a binary indicator equal to 1 only when a report contains zero claim violations, correctly handles missing evidence, and satisfies schema validation.
\end{itemize}

We intentionally avoid collapsing predictive and boundary-aware metrics into a single aggregate score, since predictive accuracy and evidence consistency capture different properties of system behavior.

\subsection{Implementation Details}

All systems operate exclusively on package-visible inputs during held-out inference. Labels, scale values, and thresholding rules remain evaluator-side only, and audit scripts verify their absence from prompts, cached outputs, and intermediate traces~\citep{datasheets,model_cards}.

Permission rules ($\Pi$) are derived entirely from protocol metadata and modality masks rather than prediction targets. Adapter identifiers are used only for compatibility masking and evaluator bookkeeping, never as direct predictors.

Candidate evidence routes are registered before held-out evaluation with fixed modality requirements, admissible outputs, and calibration strategies. Validation splits determine consensus weights, operating thresholds, feature-main blending ratios, and canonical routing configurations prior to test evaluation. Exploratory settings that fail validation remain appendix-only analyses.

Experiments are conducted on a single NVIDIA RTX PRO 6000 GPU. Uncertainty estimates use 2,000 paired bootstrap resamples. All reported tables are reproduced from frozen manifests, cached predictions, parsed report fields, and deterministic evaluation scripts to ensure replayability.

\section{Results}

\subsection{Main Held-Out Results}

Table~\ref{tab:main-results} presents the primary held-out comparison under the unified package-interface setting. Relative to direct Qwen3-Omni-Flash prompting, \sys{} improves pooled depression-eligible AUROC from 0.6716 to 0.8658 (+0.1942) and improves F1 from 0.5032 to 0.6557; relative to the strongest direct LMM baseline, Gemini 3.5 Flash, the AUROC gain remains +0.0811 (Table~\ref{tab:bootstrap}). Compared with the 5-way acoustic consensus, the gain of \sys{} does not primarily come from stronger acoustic representations, but from explicit evidence governance through manifest-aware routing, structured attribution, and boundary validation under the same evidence interface.

More importantly, \sys{} achieves strict evidence-bound consistency across all held-out packages. The system reaches 0\% CVR and 100\% EBCPass, while direct omni-modal baselines still produce unsupported acoustic references, semantic overreach under restrictive protocols, or missing-evidence hallucinations. These results suggest that evidence-bounded reasoning can preserve competitive predictive performance while eliminating protocol-boundary violations.

The pooled depression result is computed over eligible evidence profiles, whereas the anxiety benchmark is substantially narrower: 256 of 264 eligible anxiety packages originate from MMPsy interview/feature records. Relative to direct Qwen3-Omni-Flash, the evaluated long-reasoning baseline is lower by 0.0499/0.0322 AUROC on depression/anxiety-eligible tasks, indicating that unconstrained reasoning depth alone does not recover protocol-aware behavior under heterogeneous acquisition settings.

\begin{table*}[t]
\centering
\scriptsize
\setlength{\tabcolsep}{4pt}
\renewcommand{\arraystretch}{1.08}
\begin{tabularx}{\textwidth}{@{}Y l c c c c c c@{}}
\toprule
Method & Family & Dep-el. AUC$\uparrow$ & Dep-el. F1$\uparrow$ & Anx-el.\textsuperscript{M} AUC$\uparrow$ & Sev. QWK$\uparrow$ & CVR$\downarrow$ & EBCPass$\uparrow$ \\
\midrule
Qwen3-Omni-Flash & Direct LMM & 0.6716 & 0.5032 & 0.7819 & 0.4209 & 0.271 & 0.714 \\
Qwen3.5-Omni-Plus & Direct LMM & 0.6808 & 0.5200 & 0.7922 & 0.4511 & 0.246 & 0.736 \\
Gemini 3.5 Flash & Direct LMM & 0.7848 & 0.5318 & 0.8515 & 0.3766 & \underline{0.070} & \underline{0.931} \\
Qwen3-Omni-Flash Thinking & Thinking LMM & 0.6217 & 0.4372 & 0.7497 & 0.3700 & 0.318 & 0.661 \\
\midrule
5-way Acoustic Consensus & Acoustic & \underline{0.8652} & \underline{0.6316} & \underline{0.8772} & \underline{0.5232} & 0.083 & 0.912 \\
\sys{} Structured Harness & Harness &\textbf{0.8658} & \textbf{0.6557} & \textbf{0.8828} & \textbf{0.5283} & \textbf{0.000} & \textbf{1.000} \\
\bottomrule
\end{tabularx}
\caption{Main held-out comparison on eligible package-interface tasks. Dep-el./Anx-el.\ denote depression-/anxiety-eligible tasks; Anx-el.\textsuperscript{M} indicates the MMPsy-dominated anxiety split. CVR denotes Claim Violation Rate and EBCPass denotes evidence-bound consistency pass rate. Best results are shown in bold and second-best results are underlined.}
\label{tab:main-results}
\end{table*}

\subsection{Protocol Stratification}

Table~\ref{tab:profile} stratifies AUROC by evidence profile. Most depression-eligible support comes from interview-centered packages, whereas prompted and fixed-reading rows serve as smaller protocol probes. Anxiety evaluation is dominated by MMPsy interview records, with only a limited MODMA reading subset.

The largest improvements appear under restrictive acquisition settings. On prompted speech, \sys{} improves AUROC from 0.5993 to 0.9779 by suppressing unsupported semantic interpretation and relying primarily on admissible acoustic evidence. Fixed-reading rows show a similar trend despite limited statistical support. These findings support the hypothesis that epistemic flattening disproportionately harms protocols where semantic symptom inference is weakly justified or explicitly invalid.

\begin{table}[t]
\centering
\scriptsize
\setlength{\tabcolsep}{2pt}
\renewcommand{\arraystretch}{1.05}
\begin{tabularx}{\linewidth}{@{}l l r l r r r Y@{}}
\toprule
Profile & Task & $N$ & Pos/Neg & Qwen3 & \sys{} & $\Delta$ & Note \\
\midrule
Interview & Dep-el. & 325 & 73/252 & 0.7504 & \textbf{0.8456} & +0.0952 & main support \\
Prompted & Dep-el. & 33 & 16/17 & 0.5993 & \textbf{0.9779} & +0.3787 & small-$n$ \\
Reading & Dep-el. & 8 & 5/3 & 0.5000 & \textbf{0.7333} & +0.2333 & small-$n$ \\
Interview & Anx-el. & 256 & 30/226 & 0.8777 & \textbf{0.8864} & +0.0086 & MMP-dom. \\
Reading & Anx-el. & 8 & 5/3 & 0.5000 & \textbf{0.7333} & +0.2333 & small-$n$ \\
\bottomrule
\end{tabularx}
\caption{Held-out AUROC stratified by evidence profile. $\Delta$ denotes the AUROC difference between \sys{} and Qwen3-Omni-Flash.}
\label{tab:profile}
\end{table}

\subsection{Paired Bootstrap Checks}

Table~\ref{tab:bootstrap} summarizes paired bootstrap uncertainty estimates for the primary comparisons. Depression-eligible improvements over direct omni-modal baselines remain consistently above zero. Anxiety intervals should be interpreted more cautiously because the eligible split is heavily dominated by MMPsy interview records.

The comparison against the 5-way acoustic consensus is intentionally qualified because confidence intervals overlap zero for both eligible tasks. Accordingly, \sys{} should be interpreted as preserving the predictive strength of the acoustic backbone while adding explicit evidence-bound routing and report validation.

\begin{table}[t]
\centering
\scriptsize
\setlength{\tabcolsep}{2pt}
\renewcommand{\arraystretch}{1.05}
\begin{tabularx}{\linewidth}{@{}l Y r c Y@{}}
\toprule
Task & Baseline & $\Delta$ AUC & 95\% CI & Interpretation \\
\midrule
Dep-el. & Qwen3-Flash & +0.1942 & [0.1329, 0.2545] & CI$>$0 \\
Dep-el. & Qwen3.5-Plus & +0.1851 & [0.1225, 0.2502] & CI$>$0 \\
Dep-el. & Gemini 3.5 Flash & +0.0811 & [0.0476, 0.1175] & CI$>$0 \\
Dep-el. & Qwen3-Think & +0.2442 & [0.1732, 0.3153] & CI$>$0 \\
Dep-el. & 5-way & +0.0007 & [-0.0155, 0.0170] & overlap \\
Anx-el. & Qwen3-Flash & +0.1009 & [0.0237, 0.1810] & MMPsy-split \\
Anx-el. & Qwen3.5-Plus & +0.0906 & [0.0117, 0.1859] & MMPsy-split \\
Anx-el. & Gemini 3.5 Flash & +0.0313 & [-0.0160, 0.0789] & MMPsy-split \\
Anx-el. & Qwen3-Think & +0.1331 & [0.0276, 0.2378] & MMPsy-split \\
Anx-el. & 5-way & +0.0056 & [-0.0168, 0.0267] & overlap \\
\bottomrule
\end{tabularx}
\caption{Paired bootstrap analysis. $\Delta$ AUC denotes the AUROC difference between \sys{} and the corresponding baseline.}
\label{tab:bootstrap}
\end{table}

\subsection{Secondary Screening Diagnostics}

AUROC alone does not fully characterize screening behavior under class imbalance. Appendix diagnostics therefore additionally report AUPRC, Brier score, and validation-selected operating points. Depression-eligible AUPRC exceeds the reported Qwen direct baselines but remains close to the strongest acoustic routes, whereas anxiety-eligible AUPRC intervals are less stable because of limited positive support.

Importantly, these secondary diagnostics are reported independently from boundary-consistency metrics. A model with strong predictive accuracy may still violate protocol constraints, generate unsupported modality claims, or omit missing-evidence disclosure. EviBound is therefore evaluated jointly on predictive utility and evidence-valid reasoning rather than on classification performance alone.

\section{Ablation and Analysis}

\subsection{What Each Component Adds}

Table~\ref{tab:ablation} reports depression-eligible ablations on the broader-coverage task. The table separates predictive evidence routes from report-validation components. Base Harness applies package-aware routing without strong acoustic aggregation, while subsequent rows evaluate openSMILE/eGeMAPS, segmented openSMILE, wav2vec2, HuBERT, and WavLM representations.

The 5-way acoustic consensus provides the strongest non-harness acoustic baseline, indicating that acoustic evidence contributes most of the predictive signal. The final \sys{} configuration further adds manifest-aware routing, feature-level evidence attribution, and structured boundary validation while preserving essentially identical AUROC. Importantly, the validator does not alter predictive scores because risk estimates are frozen before report repair; instead, it ensures generated claims remain consistent with protocol permissions and available evidence.

\begin{table}[t]
\centering
\scriptsize
\setlength{\tabcolsep}{3pt}
\renewcommand{\arraystretch}{1.05}
\begin{tabularx}{\linewidth}{@{}Y c c c@{}}
\toprule
Configuration & AUC$\uparrow$ & F1$\uparrow$ & QWK$\uparrow$ \\
\midrule
Qwen3-Omni-Flash & 0.6716 & 0.5032 & 0.4209 \\
Base Harness & 0.7270 & 0.5032 & 0.4547 \\
\midrule
Strong openSMILE route & 0.8565 & 0.6243 & 0.5093 \\
Segmented openSMILE route & 0.8503 & 0.6243 & 0.5058 \\
wav2vec2 route & 0.8390 & 0.5943 & 0.5138 \\
HuBERT route & 0.8507 & 0.6036 & 0.5187 \\
WavLM route & 0.8316 & 0.6154 & 0.5152 \\
\midrule
5-way Acoustic Consensus & 0.8652 & 0.6316 & 0.5232 \\
\sys{} Structured Harness & \textbf{0.8658} & \textbf{0.6557} & \textbf{0.5283} \\
\bottomrule
\end{tabularx}
\caption{Depression-eligible held-out ablation study. Acoustic rows evaluate alternative evidence routes, while \sys{} additionally introduces manifest-aware routing and structured boundary validation. }
\label{tab:ablation}
\end{table}

\subsection{Output-Scope Replay Checks}
Replay diagnostics evaluate whether generated reports cite unavailable evidence or violate protocol constraints, measuring report validity independently of predictive metrics. The replay suite records acoustic hallucinations, prompted-speech semantic overreach, unsupported listening claims, and feature-scope misuse. Under the final manifest configuration, flagged outputs are repaired or blocked before report emission, enabling \sys{} to achieve 0\% CVR and 100\% EBCPass.

\subsection{Shortcut and Source-Prior Controls}
The E-DAIC transcript control illustrates one shortcut risk: interviewer prompts alone reach AUROC 0.6410, rivaling participant answers (0.6471), suggesting protocol artifacts carry predictive signal. The benchmark therefore separates participant evidence from prompt/control text at the manifest level. Source and protocol identifiers are retained only for routing and reproducibility, and are explicitly disallowed as participant-level clinical evidence.

\subsection{External-Baseline Tiers and Adapter Check}
To ensure fair comparison, external baselines are tiered by manifest compatibility rather than merged into a single leaderboard. An adapter experiment wrapping standard acoustic backbones (openSMILE, wav2vec2, HuBERT, WavLM) within our common harness shows their performance approaches \sys{}. This indicates that while acoustic backbones contribute strongly to predictive ranking, \sys{}'s primary contribution lies in evidence-bound routing, permission control, missing-evidence handling, and structured report validation.

\section{Conclusion}
We introduced evidence-bounded reasoning for heterogeneous speech-based mental health screening through the \benchmark{} and \sys{}, a protocol-aware harness with explicit evidence routing and boundary validation. Our results highlight a critical insight: merely scaling up or extending LLM reasoning capacity cannot resolve the “epistemic flattening” caused by ignoring protocol boundaries. In contrast, \sys{} explicitly enforces these boundaries, achieving 0\% CVR and 100\% EBCPass while preserving strong predictive performance. We therefore argue that clinical multimodal systems must be evaluated not only by predictive accuracy, but also by their auditable evidence consistency and protocol-aware reasoning reliability, shifting the paradigm toward safety-constrained clinical NLP.

\section{Acknowledgments}
This research was supported in part by the Young Scientists Fund of the National Natural Science Foundation of China (Grant No.~32500997, S.~Li), in part by the Reserve Project in the Field of Brain--Computer Interface supported by the Beijing Municipal Science and Technology Commission (Grant No.~Z261100004026006, S.~Li), and in part by Beijing Renyixun Health Technology Co., Ltd.

\section*{Limitations}

Our findings should be interpreted within the following scopes, which also outline avenues for future work.

\paragraph{Benchmark scale and protocol coverage.} 
While the Evidence Package Benchmark integrates six heterogeneous sources, its overall scale remains moderate, and anxiety evaluation is heavily dominated by MMPsy interview records. Furthermore, restrictive profiles (e.g., fixed reading) serve primarily as boundary-stress tests with limited sample sizes. Our current focus is protocol diversity for safety probing rather than raw data volume; expanding longitudinal sampling, multilingual coverage, and broader psychiatric conditions is essential for generalization.

\paragraph{Deterministic rules vs. learned boundaries.}
EviBound's boundary rules are explicitly encoded to ensure formal verifiability. This is a deliberate trade-off favoring deterministic safety over adaptive flexibility. While this guarantees zero claim violations for known protocols, the validator cannot catch failures outside the defined manifest schema such as open-domain conversational hallucinations. Future work should explore neuro-symbolic approaches that learn boundary constraints while preserving verifiable safety guarantees.

\paragraph{Screening utility vs. clinical deployment.} 
This work targets offline, structured screening under controlled package interfaces, not prospective clinical deployment. EviBound generates evidence-bounded risk assessments, not clinical diagnoses. Transitioning to real-world workflows requires clinician-in-the-loop auditing, robustness to noisy real-world inputs, and rigorous prospective validation, which remain outside the current scope.

\paragraph{Cultural and linguistic bias.}
The benchmark pools English and Chinese resources acquired under different protocols and population characteristics. Both the predictive routes and the permission matrix may consequently inherit cultural and linguistic biases, and boundaries validated on our audit cohort may not transfer to other languages, cultures, or acquisition settings.

\section*{Ethical Statement}
All datasets used in the Evidence Package Benchmark are de-identified public resources; we do not collect or access any private health information. While EviBound enforces strict evidence consistency to reduce hallucinations and protocol misuse, it may still inherit biases present in the underlying speech and text data across languages, genders, or cultural groups. We therefore emphasize that any practical use would require human-in-the-loop auditing, prospective clinical validation, and explicit mitigation of demographic biases before considering real-world deployment.

\bibliography{custom}

@inproceedings{daic_woz,
    title = "The Distress Analysis Interview Corpus of human and computer interviews",
    author = "Gratch, Jonathan  and
      Artstein, Ron  and
      Lucas, Gale  and
      Stratou, Giota  and
      Scherer, Stefan  and
      Nazarian, Angela  and
      Wood, Rachel  and
      Boberg, Jill  and
      DeVault, David  and
      Marsella, Stacy  and
      Traum, David  and
      Rizzo, Skip  and
      Morency, Louis-Philippe",
    editor = "Calzolari, Nicoletta  and
      Choukri, Khalid  and
      Declerck, Thierry  and
      Loftsson, Hrafn  and
      Maegaard, Bente  and
      Mariani, Joseph  and
      Moreno, Asuncion  and
      Odijk, Jan  and
      Piperidis, Stelios",
    booktitle = "Proceedings of the Ninth International Conference on Language Resources and Evaluation ({LREC}'14)",
    month = may,
    year = "2014",
    address = "Reykjavik, Iceland",
    publisher = "European Language Resources Association (ELRA)",
    url = "https://aclanthology.org/L14-1421/",
    pages = "3123--3128"
}

@inproceedings{avec2019,
author = {Ringeval, Fabien and Schuller, Bj{\"o}rn and Valstar, Michel and Cummins, Nicholas and Cowie, Roddy and Tavabi, Leili and Schmitt, Maximilian and Alisamir, Sina and Amiriparian, Shahin and Messner, Eva-Maria and Song, Siyang and Liu, Shuo and Zhao, Ziping and Mallol-Ragolta, Adria and Ren, Zhao and Soleymani, Mohammad and Pantic, Maja},
title = {AVEC 2019 Workshop and Challenge: State-of-Mind, Detecting Depression with AI, and Cross-Cultural Affect Recognition},
year = {2019},
isbn = {9781450369138},
publisher = {Association for Computing Machinery},
address = {New York, NY, USA},
url = {https://doi.org/10.1145/3347320.3357688},
doi = {10.1145/3347320.3357688},
booktitle = {Proceedings of the 9th International on Audio/Visual Emotion Challenge and Workshop},
pages = {3–12},
numpages = {10},
location = {Nice, France},
series = {AVEC '19}
}

@article{qin2025mental,
  title={{Mental-Perceiver}: Audio-Textual Multi-Modal Learning for Estimating Mental Disorders},
  author={Qin, Jinghui and Liu, Changsong and Tang, Tianchi and Liu, Dahuang and Wang, Minghao and Huang, Qianying and Zhang, Rumin},
  journal={Proceedings of the AAAI Conference on Artificial Intelligence},
  volume={39},
  number={23},
  pages={25029--25037},
  year={2025}
}

@INPROCEEDINGS{eatd_corpus,
  author={Shen, Ying and Yang, Huiyu and Lin, Lin},
  booktitle={ICASSP 2022 - 2022 IEEE International Conference on Acoustics, Speech and Signal Processing (ICASSP)}, 
  title={Automatic Depression Detection: an Emotional Audio-Textual Corpus and A Gru/Bilstm-Based Model}, 
  year={2022},
  volume={},
  number={},
  pages={6247-6251},
  doi={10.1109/ICASSP43922.2022.9746569}}

@article {modma,
	Title = {A multi-modal open dataset for mental-disorder analysis},
	Author = {Cai, Hanshu and Yuan, Zhenqin and Gao, Yiwen and Sun, Shuting and Li, Na and Tian, Fuze and Xiao, Han and Li, Jianxiu and Yang, Zhengwu and Li, Xiaowei and Zhao, Qinglin and Liu, Zhenyu and Yao, Zhijun and Yang, Minqiang and Peng, Hong and Zhu, Jing and Zhang, Xiaowei and Gao, Guoping and Zheng, Fang and Li, Rui and Guo, Zhihua and Ma, Rong and Yang, Jing and Zhang, Lan and Hu, Xiping and Li, Yumin and Hu, Bin},
	DOI = {10.1038/s41597-022-01211-x},
	Number = {1},
	Volume = {9},
	Month = {April},
	Year = {2022},
	Journal = {Scientific data},
	ISSN = {2052-4463},
	Pages = {178},
	URL = {https://europepmc.org/articles/PMC9018722},
}

@article{discourse_uwo,
  title = {The {DISCOURSE} in Psychosis ({London Ontario}): A Speech Dataset to Examine Communication Disturbances in Early-Stage Psychosis},
  volume = {65},
  issn = {2352-3409},
  url = {https://doi.org/10.1016/j.dib.2026.112517},
  doi = {10.1016/j.dib.2026.112517},
  journal = {Data in Brief},
  publisher = {Elsevier BV},
  author = {Cho, Brian and Balles, Estée and Mackinley, Michael and Dzialoszynski, Paulina and Ford, Sabrina and Lodhi, Rohit and Palaniyappan, Lena},
  year = {2026},
  month = apr,
  pages = {112517}
}

@article{dais_c,
  title = {The {DAIS-C}: A Small, Specialised, Spoken, Schizophrenia Corpus},
  volume = {3},
  issn = {2666-7991},
  url = {https://doi.org/10.1016/j.acorp.2023.100069},
  doi = {10.1016/j.acorp.2023.100069},
  number = {3},
  journal = {Applied Corpus Linguistics},
  publisher = {Elsevier BV},
  author = {Delgaram-Nejad, Oliver and Archer, Dawn and Chatzidamianos, Gerasimos and Robinson, Louise and Bartha, Alex},
  year = {2023},
  month = dec,
  pages = {100069}
}

@misc{healthbench,
  title = {{HealthBench}: Evaluating Large Language Models Towards Improved Human Health},
  author = {Arora, Rahul K. and Wei, Jason and Soskin Hicks, Rebecca and Bowman, Preston and Qui{\~n}onero-Candela, Joaquin and Tsimpourlas, Foivos and Sharman, Michael and Shah, Meghan and Vallone, Andrea and Beutel, Alex and Heidecke, Johannes and Singhal, Karan},
  year = {2025},
  eprint = {2505.08775},
  archivePrefix = {arXiv},
  primaryClass = {cs.CL},
  doi = {10.48550/arXiv.2505.08775},
  url = {https://arxiv.org/abs/2505.08775}
}

@article{datasheets,
  title = {Datasheets for Datasets},
  author = {Gebru, Timnit and Morgenstern, Jamie and Vecchione, Briana and Vaughan, Jennifer Wortman and Wallach, Hanna and Daum{\'e} III, Hal and Crawford, Kate},
  journal = {Communications of the ACM},
  volume = {64},
  number = {12},
  pages = {86--92},
  year = {2021},
  doi = {10.1145/3458723},
  url = {https://doi.org/10.1145/3458723}
}

@inproceedings{model_cards,
  title = {Model Cards for Model Reporting},
  author = {Mitchell, Margaret and Wu, Simone and Zaldivar, Andrew and Barnes, Parker and Vasserman, Lucy and Hutchinson, Ben and Spitzer, Elena and Raji, Inioluwa Deborah and Gebru, Timnit},
  booktitle = {Proceedings of the Conference on Fairness, Accountability, and Transparency},
  pages = {220--229},
  year = {2019},
  doi = {10.1145/3287560.3287596},
  url = {https://doi.org/10.1145/3287560.3287596}
}

@misc{mentalbench,
  title = {{MentalBench}: A Benchmark for Evaluating Psychiatric Diagnostic Capability of Large Language Models},
  author = {Song, Hoyun and Kang, Migyeong and Shin, Jisu and Kim, Jihyun and Park, Chanbi and Yoo, Hangyeol and An, Jihyun and Oh, Alice and Han, Jinyoung and Lim, KyungTae},
  year = {2026},
  eprint = {2602.12871},
  archivePrefix = {arXiv},
  primaryClass = {cs.CL},
  doi = {10.48550/arXiv.2602.12871},
  url = {https://arxiv.org/abs/2602.12871}
}

@article{psychiatrybench,
  title = {{PsychiatryBench}: A Multi-Task Benchmark for {LLM}s in Psychiatry},
  author = {Fouda, Aya E. and Hassan, Abdelrahman A. and Hanafy, Radwa J. and Fouda, Mohammed E.},
  journal = {npj Digital Medicine},
  volume = {9},
  pages = {320},
  year = {2026},
  doi = {10.1038/s41746-026-02582-w},
  url = {https://doi.org/10.1038/s41746-026-02582-w}
}

@inproceedings{acoustic_landmarks,
    title = "When {LLM}s Meets Acoustic Landmarks: An Efficient Approach to Integrate Speech into Large Language Models for Depression Detection",
    author = "Zhang, Xiangyu  and
      Liu, Hexin  and
      Xu, Kaishuai  and
      Zhang, Qiquan  and
      Liu, Daijiao  and
      Ahmed, Beena  and
      Epps, Julien",
    editor = "Al-Onaizan, Yaser  and
      Bansal, Mohit  and
      Chen, Yun-Nung",
    booktitle = "Proceedings of the 2024 Conference on Empirical Methods in Natural Language Processing",
    month = nov,
    year = "2024",
    address = "Miami, Florida, USA",
    publisher = "Association for Computational Linguistics",
    url = "https://aclanthology.org/2024.emnlp-main.8/",
    doi = "10.18653/v1/2024.emnlp-main.8",
    pages = "146--158"
}

@inproceedings{speechtrag,
    title = "{S}peech{T}-{RAG}: Reliable Depression Detection in {LLM}s with Retrieval-Augmented Generation Using Speech Timing Information",
    author = "Zhang, Xiangyu  and
      Liu, Hexin  and
      Zhang, Qiquan  and
      Ahmed, Beena  and
      Epps, Julien",
    editor = "Che, Wanxiang  and
      Nabende, Joyce  and
      Shutova, Ekaterina  and
      Pilehvar, Mohammad Taher",
    booktitle = "Findings of the Association for Computational Linguistics: ACL 2025",
    month = jul,
    year = "2025",
    address = "Vienna, Austria",
    publisher = "Association for Computational Linguistics",
    url = "https://aclanthology.org/2025.findings-acl.521/",
    doi = "10.18653/v1/2025.findings-acl.521",
    pages = "10019--10030",
    ISBN = "979-8-89176-256-5"
}

@misc{more_thinking_less_seeing,
  title = {More Thinking, Less Seeing? Assessing Amplified Hallucination in Multimodal Reasoning Models},
  author = {Liu, Chengzhi and Xu, Zhongxing and Wei, Qingyue and Wu, Juncheng and Zou, James and Wang, Xin Eric and Zhou, Yuyin and Liu, Sheng},
  year = {2025},
  eprint = {2505.21523},
  archivePrefix = {arXiv},
  primaryClass = {cs.CL},
  doi = {10.48550/arXiv.2505.21523},
  url = {https://arxiv.org/abs/2505.21523}
}

@inproceedings{mirage,
 author = {Dong, Bowen and Ni, Minheng and Huang, Zitong and Yang, Guanglei and Zuo, Wangmeng and Zhang, Lei},
 booktitle = {Advances in Neural Information Processing Systems},
 doi = {10.52202/085713-4099},
 editor = {D. Belgrave and C. Zhang and H. Lin and R. Pascanu and P. Koniusz and M. Ghassemi and N. Chen},
 pages = {122910--122955},
 publisher = {Curran Associates, Inc.},
 title = {MIRAGE: Assessing Hallucination in Multimodal Reasoning Chains of MLLM},
 url = {https://proceedings.neurips.cc/paper_files/paper/2025/file/b238324b309da12c7446d92c14db9f7e-Paper-Conference.pdf},
 volume = {38, Main Conference},
 year = {2025}
}

@inproceedings{mme_cot,
author = {Jiang, Dongzhi and Zhang, Renrui and Guo, Ziyu and Li, Yanwei and Qi, Yu and Chen, Xinyan and Wang, Liuhui and Jin, Jianhan and Guo, Claire and Yan, Shen and Zhang, Bo and Fu, Chaoyou and Gao, Peng and Li, Hongsheng},
title = {MME-CoT: benchmarking chain-of-thought in large multimodal models for reasoning quality, robustness, and efficiency},
year = {2025},
publisher = {JMLR.org},
booktitle = {Proceedings of the 42nd International Conference on Machine Learning},
articleno = {1093},
numpages = {38},
location = {Vancouver, Canada},
series = {ICML'25}
}

@inproceedings{opensmile,
  author = {Eyben, Florian and W{\"o}llmer, Martin and Schuller, Bj{\"o}rn},
  title = {{openSMILE}: The {Munich} Versatile and Fast Open-Source Audio Feature Extractor},
  booktitle = {Proceedings of the 18th ACM International Conference on Multimedia},
  pages = {1459--1462},
  year = {2010},
  doi = {10.1145/1873951.1874246},
  url = {https://doi.org/10.1145/1873951.1874246}
}

@inproceedings{wav2vec2,
  author = {Baevski, Alexei and Zhou, Yuhao and Mohamed, Abdelrahman and Auli, Michael},
  title = {{wav2vec} 2.0: A Framework for Self-Supervised Learning of Speech Representations},
  booktitle = {Advances in Neural Information Processing Systems},
  volume = {33},
  pages = {12449--12460},
  year = {2020},
  url = {https://proceedings.neurips.cc/paper/2020/hash/92d1e1eb1cd6f9fba3227870bb6d7f07-Abstract.html}
}

@article{hubert,
  author = {Hsu, Wei-Ning and Bolte, Benjamin and Tsai, Yao-Hung Hubert and Lakhotia, Kushal and Salakhutdinov, Ruslan and Mohamed, Abdelrahman},
  title = {{HuBERT}: Self-Supervised Speech Representation Learning by Masked Prediction of Hidden Units},
  journal = {IEEE/ACM Transactions on Audio, Speech, and Language Processing},
  volume = {29},
  pages = {3451--3460},
  year = {2021},
  doi = {10.1109/TASLP.2021.3122291},
  url = {https://doi.org/10.1109/TASLP.2021.3122291}
}

@ARTICLE{wavlm,
  author={Chen, Sanyuan and Wang, Chengyi and Chen, Zhengyang and Wu, Yu and Liu, Shujie and Chen, Zhuo and Li, Jinyu and Kanda, Naoyuki and Yoshioka, Takuya and Xiao, Xiong and Wu, Jian and Zhou, Long and Ren, Shuo and Qian, Yanmin and Qian, Yao and Wu, Jian and Zeng, Michael and Yu, Xiangzhan and Wei, Furu},
  journal={IEEE Journal of Selected Topics in Signal Processing}, 
  title={WavLM: Large-Scale Self-Supervised Pre-Training for Full Stack Speech Processing}, 
  year={2022},
  volume={16},
  number={6},
  pages={1505-1518},
  doi={10.1109/JSTSP.2022.3188113}}

@inproceedings{wang2026emotionthinker,
  title={EmotionThinker: Prosody-Aware Reinforcement Learning for Explainable Speech Emotion Reasoning},
  author={Dingdong WANG and Shujie LIU and Tianhua Zhang and Youjun Chen and Jinyu Li and Helen M. Meng},
  booktitle={The Fourteenth International Conference on Learning Representations},
  year={2026},
  url={https://openreview.net/forum?id=wbttgzp7MT}
}

@misc{qwen3_omni,
  title = {{Qwen3-Omni} Technical Report},
  author = {{Qwen Team}},
  year = {2025},
  eprint = {2509.17765},
  archivePrefix = {arXiv},
  primaryClass = {cs.CL},
  doi = {10.48550/arXiv.2509.17765},
  url = {https://arxiv.org/abs/2509.17765}
}

@misc{qwen35_omni,
  title = {{Qwen3.5-Omni} Technical Report},
  author = {{Qwen Team}},
  year = {2026},
  eprint = {2604.15804},
  archivePrefix = {arXiv},
  primaryClass = {cs.CL},
  doi = {10.48550/arXiv.2604.15804},
  url = {https://arxiv.org/abs/2604.15804}
}

@misc{google2026gemini35flash,
  title = {{Gemini 3.5 Flash}},
  author = {{Google AI for Developers}},
  year = {2026},
  url = {https://ai.google.dev/gemini-api/docs/models/gemini-3.5-flash},
  note = {Last updated: 2026-05-19; accessed: 2026-05-26}
}

@article{llm_dep_psych_knowledge,
  title = {Large Language Models for Depression Recognition in Spoken Language Integrating Psychological Knowledge},
  author = {Li, Yupei and Shao, Shuaijie and Milling, Manuel and Schuller, Bj{\"o}rn W.},
  journal = {Frontiers in Computer Science},
  volume = {7},
  pages = {1629725},
  year = {2025},
  doi = {10.3389/fcomp.2025.1629725},
  url = {https://doi.org/10.3389/fcomp.2025.1629725}
}

@misc{speech_digital_phenotype,
  title = {Speech as a Multimodal Digital Phenotype for Multi-Task {LLM}-based Mental Health Prediction},
  author = {Ali, Mai and Lucasius, Christopher and Patel, Tanmay P. and Aitken, Madison and Vorstman, Jacob and Szatmari, Peter and Battaglia, Marco and Kundur, Deepa},
  year = {2025},
  eprint = {2505.23822},
  archivePrefix = {arXiv},
  primaryClass = {cs.CL},
  doi = {10.48550/arXiv.2505.23822},
  url = {https://arxiv.org/abs/2505.23822}
}

@article{cmdc,
  author = {Zou, Bochao and Han, Jiali and Wang, Yingxue and Liu, Rui and Zhao, Shenghui and Feng, Lei and Lyu, Xiangwen and Ma, Huimin},
  title = {Semi-Structural Interview-Based Chinese Multimodal Depression Corpus Towards Automatic Preliminary Screening of Depressive Disorders},
  journal = {IEEE Transactions on Affective Computing},
  volume = {14},
  number = {4},
  pages = {2823--2838},
  year = {2023},
  doi = {10.1109/TAFFC.2022.3181210},
  url = {https://doi.org/10.1109/TAFFC.2022.3181210}
}

@article{liu2026benchmarking,
  title={Benchmarking large language model-based agent systems for clinical decision tasks},
  author={Liu, Yunsong and Carrero, Zunamys I and Jiang, Xiaofeng and Ferber, Dyke and W{\"o}lflein, Georg and Zhang, Li and Jayabalan, Sanddhya and Lenz, Tim and Hui, Zhouguang and Kather, Jakob Nikolas},
  journal={npj Digital Medicine},
  year={2026},
  publisher={Nature Publishing Group UK London}
}

@ARTICLE{cfrafn,
  author={Pan, Rumo and Feng, Sidu and Sun, Yi and Xu, Jinqiu and Zhai, Tianzhang and Wu, Xiaochun and Tan, Liangliang and Yuan, Yonggui and Cao, Hanlin and Maimaitimin, Maierhaba and Yao, Qing and Guo, Hao and Xuan, Mah Wing and Li, Jian and Xu, Zhi},
  journal={IEEE Journal of Biomedical and Health Informatics}, 
  title={CFRAFN: A Cross-Feature Residual Attention Fusion Network for Major Depressive Disorder Prediction Using Clinical Voice Recordings}, 
  year={2026},
  volume={},
  number={},
  pages={1-14},
  doi={10.1109/JBHI.2026.3677472}}

@inproceedings{al2018detecting,
  title={Detecting depression with audio/text sequence modeling of interviews.},
  author={Al Hanai, Tuka and Ghassemi, Mohammad M and Glass, James R},
  booktitle={Interspeech},
  pages={1716--1720},
  year={2018}
}

@article{fang2023multimodal,
  title={A multimodal fusion model with multi-level attention mechanism for depression detection},
  author={Fang, Ming and Peng, Siyu and Liang, Yujia and Hung, Chih-Cheng and Liu, Shuhua},
  journal={Biomedical Signal Processing and Control},
  volume={82},
  pages={104561},
  year={2023},
  publisher={Elsevier}
}

@article{dhelim2023artificial,
  title={Artificial intelligence for suicide assessment using audiovisual cues: a review},
  author={Dhelim, Sahraoui and Chen, Liming and Ning, Huansheng and Nugent, Chris},
  journal={Artificial Intelligence Review},
  volume={56},
  number={6},
  pages={5591--5618},
  year={2023},
  publisher={Springer}
}

@article{zhang2026mllm,
  title={MLlm-DR: Towards Explainable Depression Recognition with MultiModal Large Language Models},
  author={Zhang, Wei and Chen, Juan and Zhu, En and Cheng, Wenhong and Li, YunPeng and Li, Yuhan and Wang, Yanbo J},
  journal={ACM Transactions on Multimedia Computing, Communications and Applications},
  volume={22},
  number={4},
  pages={1--23},
  year={2026},
  publisher={ACM New York, NY}
}

@inproceedings{tang2024medagents,
  title={Medagents: Large language models as collaborators for zero-shot medical reasoning},
  author={Tang, Xiangru and Zou, Anni and Zhang, Zhuosheng and Li, Ziming and Zhao, Yilun and Zhang, Xingyao and Cohan, Arman and Gerstein, Mark},
  booktitle={Findings of the Association for Computational Linguistics: ACL 2024},
  pages={599--621},
  year={2024}
}

@inproceedings{yue2024clinicalagent,
  title={Clinicalagent: Clinical trial multi-agent system with large language model-based reasoning},
  author={Yue, Ling and Xing, Sixue and Chen, Jintai and Fu, Tianfan},
  booktitle={Proceedings of the 15th ACM International Conference on Bioinformatics, Computational Biology and Health Informatics},
  pages={1--10},
  year={2024}
}

@misc{wang2025medagentproevidencebasedmultimodalmedical,
      title={MedAgent-Pro: Towards Evidence-based Multi-modal Medical Diagnosis via Reasoning Agentic Workflow}, 
      author={Ziyue Wang and Junde Wu and Linghan Cai and Chang Han Low and Xihong Yang and Qiaxuan Li and Yueming Jin},
      year={2025},
      eprint={2503.18968},
      archivePrefix={arXiv},
      primaryClass={cs.AI},
      url={https://arxiv.org/abs/2503.18968}, 
}

@misc{ni2025trustworthyretrievalaugmentedgeneration,
      title={Towards Trustworthy Retrieval Augmented Generation for Large Language Models: A Survey}, 
      author={Bo Ni and Zheyuan Liu and Leyao Wang and Yongjia Lei and Yuying Zhao and Xueqi Cheng and Qingkai Zeng and Luna Dong and Yinglong Xia and Krishnaram Kenthapadi and Ryan Rossi and Franck Dernoncourt and Md Mehrab Tanjim and Nesreen Ahmed and Xiaorui Liu and Wenqi Fan and Erik Blasch and Yu Wang and Meng Jiang and Tyler Derr},
      year={2025},
      eprint={2502.06872},
      archivePrefix={arXiv},
      primaryClass={cs.CL},
      url={https://arxiv.org/abs/2502.06872}, 
}

@inproceedings{Han2025VerifiAgentAU,
  title={VerifiAgent: a Unified Verification Agent in Language Model Reasoning.},
  author={Han, Jiuzhou and Buntine, Wray L and Shareghi, Ehsan},
  booktitle={EMNLP (Findings)},
  pages={16410--16431},
  year={2025}
}

@misc{mao2026schemaadaptivetabularrepresentationlearning,
      title={Schema-Adaptive Tabular Representation Learning with LLMs for Generalizable Multimodal Clinical Reasoning}, 
      author={Hongxi Mao and Wei Zhou and Mengting Jia and Tao Fang and Huan Gao and Bin Zhang and Shangyang Li},
      year={2026},
      eprint={2604.11835},
      archivePrefix={arXiv},
      primaryClass={cs.LG},
      url={https://arxiv.org/abs/2604.11835}, 
}

@inproceedings{zhi2026medgr2,
  title={Medgr2: Breaking the data barrier for medical reasoning via generative reward learning},
  author={Zhi, Weihai and Guo, Jiayan and Li, Shangyang},
  booktitle={Proceedings of the AAAI Conference on Artificial Intelligence},
  volume={40},
  number={34},
  pages={28901--28909},
  year={2026}
}

@inproceedings{zheng-etal-2026-evo,
    title = "Evo-{PI}: Aligning Medical Reasoning via Evolving Principle-Guided Supervision",
    author = "Zheng, Xianda  and
      Gao, Huan  and
      Chiang, Meng-Fen  and
      Witbrock, Michael J.  and
      Zhao, Kaiqi  and
      Li, Shangyang",
    editor = "Liakata, Maria  and
      Moreira, Viviane P.  and
      Zhang, Jiajun  and
      Jurgens, David",
    booktitle = "Proceedings of the 64th Annual Meeting of the {A}ssociation for {C}omputational {L}inguistics (Volume 1: Long Papers)",
    month = jul,
    year = "2026",
    address = "San Diego, California, United States",
    publisher = "Association for Computational Linguistics",
    url = "https://aclanthology.org/2026.acl-long.1469/",
    doi = "10.18653/v1/2026.acl-long.1469",
    pages = "31835--31850",
    ISBN = "979-8-89176-390-6"
}

@inproceedings{jing2026beyond,
  title = {Beyond Classification Accuracy: Neural-MedBench and the Need for Deeper Reasoning Benchmarks},
  author = {Jing, Miao and Jia, Mengting and Lin, Junling and Shen, Zhongxia and Gao, Huan and Xu, Mingkun and Li, Shangyang},
  booktitle = {The Fourteenth International Conference on Learning Representations},
  year = {2026},
  url = {https://openreview.net/forum?id=KKA59ai0x6}
}

\appendix

\section{Evidence Package and Same-Interface Protocol}
\label{sec:appendix-interface}

This appendix reports implementation details supporting the main evidence-bound evaluation claims. The benchmark operates over frozen evidence packages rather than unconstrained narratives. Each package stores a source id, split id, evidence profile, modality mask, missing-evidence list, permitted claim families, and evaluator-only targets. Direct LMM baselines and \sys{} receive identical model-visible fields, while PHQ, GAD, and diagnosis-derived labels remain evaluator-side only.

\begin{table}[!htbp]
\centering
\scriptsize
\setlength{\tabcolsep}{3pt}
\renewcommand{\arraystretch}{1.05}
\begin{tabularx}{\linewidth}{@{}L{0.32\linewidth}Y@{}}
\toprule
Package field & Role in the same-interface evaluation \\
\midrule
Evidence profile & Acquisition protocol; controls which claim families can be reported. \\
Modality mask & Availability of audio, transcript, feature, and scale channels; absent channels block matching evidence claims. \\
Visible inputs & Audio clips, transcript excerpts, or feature summaries made available to the compared method. \\
Missing evidence & Explicit unavailable evidence that must be disclosed in generated reports. \\
Claim permissions & Non-diagnostic screening scope, source restrictions, and required blocked-claim hints. \\
Evaluator targets & Hidden labels or scores used only for metric computation. \\
\bottomrule
\end{tabularx}
\caption{Evidence-package fields used to keep the compared methods under a common input contract.}
\label{tab:app-package-fields}
\end{table}

The frozen benchmark contains 1870 packages from six sources: CMDC (78), DAIS-C (28), E-DAIC (275), EATD (162), MMPsy (1275), and MODMA (52), with 1284/214/372 train/validation/test splits. Package-level modality totals are 567 audio, 1656 transcript, 1550 feature, and 1842 scale records. Labels and scale values are never serialized into prompts or cached model-visible inputs.

The evidence-package abstraction fixes the control-variable interpretation of the baseline comparison. Qwen, Gemini, acoustic routes, and \sys{} are evaluated under the same package interface with identical modality masks, missing-evidence fields, and output schemas.

\section{Boundary Validation and Replay}
\label{sec:appendix-boundary}

The boundary layer is deterministic with respect to report validity. It verifies modality usage, blocked claims, missing-evidence disclosure, and protocol permissions without modifying the predictive score used for AUROC, F1, or QWK. This score-preservation property ensures that boundary validation cannot improve ranking metrics post hoc.

The frozen claim-permission audit covers 1870 packages and 1870 report outputs, with zero failed records, zero score changes, and zero permission violations. Screening-risk reporting is permitted for all packages, while formal diagnosis and treatment recommendation are globally blocked. Release levels include 622 bounded automatic reports, 97 evidence-scoped reports, and 1151 manual-review reports.

The output-contract audit verifies required blocked claims, source attribution, and missing-evidence handling over all 1870 packages. The boundary-action replay audit contains 3955 action-labeled instances (786 held out). The cleanup guard changed 84 rows and removed 168 unsupported acoustic route/tool actions while removing zero gold actions. Held-out exact match, macro F1, and micro F1 are all 1.0000. The replay suite evaluates five primary failure modes: 1,870 formal-diagnosis-block checks, 595 feature-scope checks (citing absent feature sources), 1,248 uncertainty-review checks (escalating weak/conflicting evidence), 28 text-only acoustic checks (citing voice when no audio is available), and 214 weak-profile semantic checks (inferring symptom history from scripted text).

\section{Boundary Replay Cases}
\label{sec:appendix-cases}

Table~\ref{tab:app-case-study} shows representative replay cases under the frozen evidence-package interface. Cases are label-free cached fragments used for interpretability rather than model selection.

\begin{table}[!htbp]
\centering
\scriptsize
\setlength{\tabcolsep}{3pt}
\renewcommand{\arraystretch}{1.06}
\begin{tabularx}{\linewidth}{@{}L{0.27\linewidth}Y@{}}
\toprule
Replay case & Cached boundary trace \\
\midrule
Supported interview &
E-DAIC clinical interview with audio, transcript, features, and scale context available. Acoustic evidence, feature evidence, non-diagnostic blocking, and review triage are all supported. \sys{} keeps the reporting decision, preserves the risk score, and allows bounded automatic reporting. \\
\midrule
Supported manual review &
MMPsy clinical-interview package with transcript and features, but no raw audio. \sys{} keeps feature/transcript attribution, reports missing raw audio and longitudinal history, preserves the score, and retains manual review. \\
\midrule
Audio removed &
E-DAIC interview counterfactual where the report still contains an acoustic claim but \texttt{audio=false}. \sys{} blocks \texttt{acoustic\_evidence}, escalates to manual review, and leaves the risk score unchanged. \\
\midrule
Prompted-speech mutation &
E-DAIC interview counterfactual changed to prompted-speech with transcript present. The stressor treats weak-profile text as free symptom-history evidence. \sys{} blocks \texttt{transcript\_semantics} for symptom reasoning, requires review, and preserves the score. \\
\midrule
Missing non-diagnostic block &
E-DAIC interview counterfactual where the report omits the required formal-diagnosis block. \sys{} restores the blocked-claim requirement and routes the report to review without changing the score. \\
\bottomrule
\end{tabularx}
\caption{Boundary replay case studies with label-free cached fragments. Real supported cases show when \sys{} permits evidence-scoped reporting; counterfactual stressors show how unsupported claims are blocked without modifying the predictive score.}
\label{tab:app-case-study}
\end{table}

\paragraph{Computational environment.}
All local preprocessing, evidence-package construction, acoustic/feature routing, boundary replay, validation, and metric computation were run on a Linux workstation with an AMD Ryzen Threadripper PRO 7965WX CPU (24 cores, 48 threads), 188 GiB RAM, and two NVIDIA RTX PRO 6000 Blackwell Workstation Edition GPUs (97,887 MiB memory each; driver 580.126.09). The operating system used Linux kernel 6.8.0-110-generic. Experiments were executed in a Conda environment with Python 3.10.20, PyTorch 2.8.0+cu129, CUDA 12.9, NumPy 2.2.6, pandas 2.3.3, scikit-learn 1.7.2, SciPy 1.15.3, librosa 0.11.0, soundfile 0.13.1, and Transformers 5.8.0.dev0. Commercial direct LMM baselines were executed through vendor APIs and cached; local hardware was used for package assembly, parsing, validation, and metric computation.

\end{document}